\documentclass[10pt,twocolumn,letterpaper]{article}

\usepackage[pagenumbers]{cvpr} 

\usepackage{microtype}

\usepackage{caption}
\usepackage{graphicx}
\usepackage{algorithm,algpseudocode}

\definecolor{cvprblue}{rgb}{0.21,0.49,0.74}
\usepackage[pagebackref,breaklinks,colorlinks,allcolors=cvprblue]{hyperref}

\def\paperID{*****} 
\def\confName{CVPR}
\def\confYear{2026}

\title{Pareto LoRA: Mitigating Modality Imbalance in Unified Multimodal Models via Pareto-Optimal Gradient Integration}

\author{Xiwen Wei\\
The University of Texas at Austin\\
{\tt\small xiwenwei@utexas.edu}
\and
Mark Nutter\\
Advanced Micro Devices, Inc.\\
{\tt\small Mark.Nutter@amd.com}
\and
Madhusudhanan Srinivasan \\
Advanced Micro Devices, Inc.\\
{\tt\small Madhu.Srinivasan2@amd.com}
\and
Radu Marculescu \\
The University of Texas at Austin\\
{\tt\small radum@utexas.edu}
}
\begin{document}
\twocolumn[{%
\renewcommand\twocolumn[1][]{#1}%

\maketitle

\begin{center}
    \includegraphics[width=0.9\textwidth]{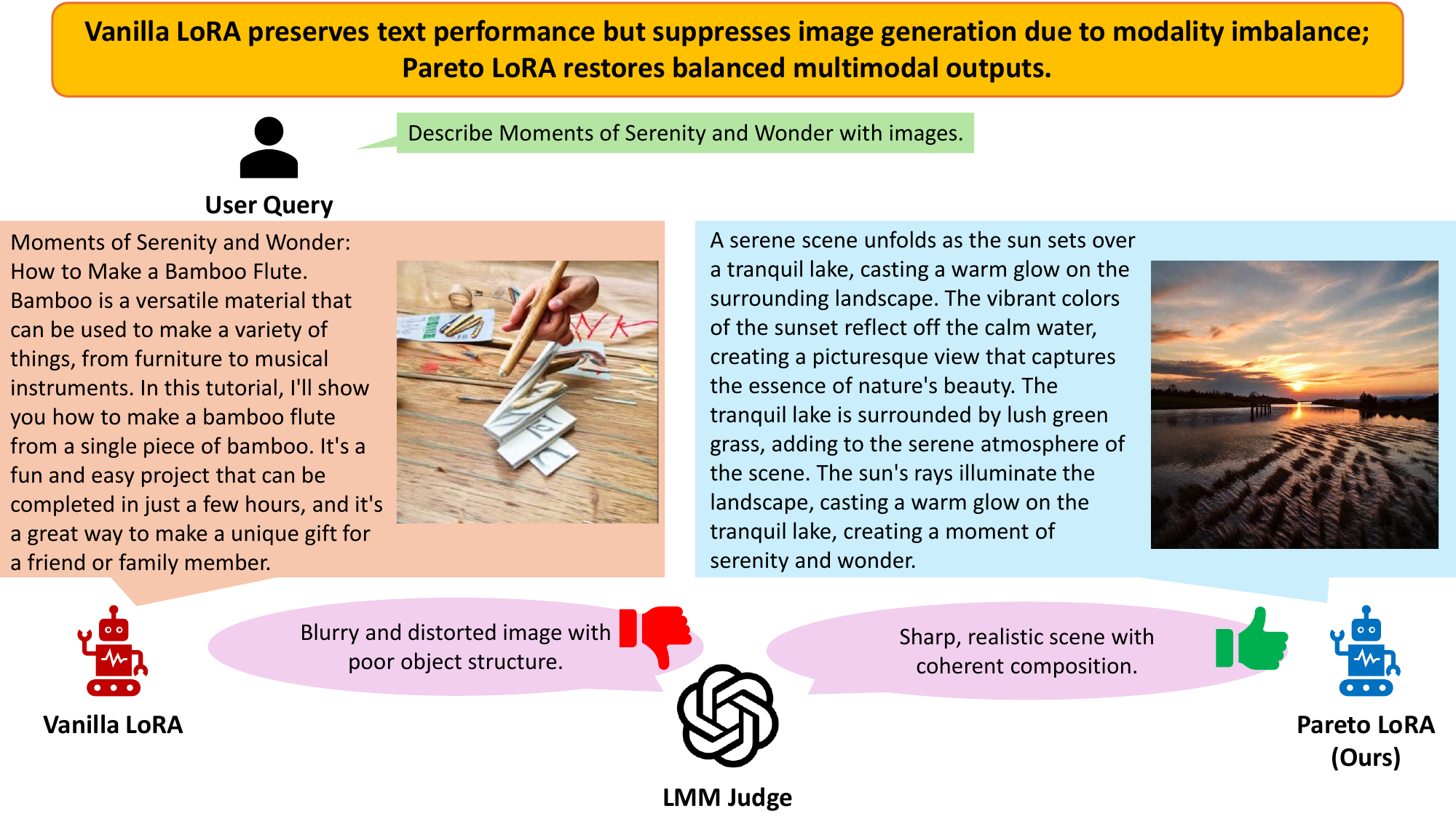}
    \captionof{figure}{Qualitative comparison on interleaved text--image generation. Vanilla LoRA preserves fluent text generation but produces blurry and distorted images due to text-dominant modality imbalance. Pareto LoRA mitigates the gradient dominance and restores coherent, high-quality image outputs while maintaining comparable text responses.}
    \label{fig:qual_task4}
\end{center}

\begin{abstract}
Unified multimodal models (UMMs) have recently emerged as a promising paradigm for integrating multimodal understanding and generation within a single autoregressive transformer.
However, during multimodal instruction tuning, these models often exhibit pronounced modality imbalance: language gradients dominate optimization, thus leading to lower image generation quality, especially under parameter-efficient fine-tuning such as LoRA.
In this work, we systematically analyze modality imbalance in LoRA-based fine-tuning of UMMs for interleaved text--image generation.
We show that vision modality performance degrades substantially more than text modality performance when compared to unimodal counterparts, and that modality-specific gradients can differ by orders of magnitude across various tasks and layers.
Motivated by this observation, we reformulate the multimodal instruction tuning as a bi-objective optimization problem and propose \emph{Pareto LoRA}, a Pareto-optimal gradient integration strategy that balances the text and image objectives by modulating the gradient direction and strength.
Experiments on the CoMM benchmark with Emu2 demonstrate that Pareto LoRA consistently improves multimodal generation balance, achieving up to 44.9\% gains in perceptual image quality over vanilla LoRA while maintaining comparable text performance.
\end{abstract}
}]    
\section{Introduction}
\label{sec:intro}


Traditional multimodal models are typically developed for either multimodal understanding
(e.g., answering questions about images) or multimodal generation
(e.g., synthesizing images from text prompts)~\cite{llava, stable_diffusion}.
Unified Multimodal Models (UMMs) seek to bridge this divide by integrating both capabilities
within a single framework.
Recent advances in UMMs~\cite{chameleon, janus, transfusion, emu3, show-o, orthus}
have demonstrated promising performance across a broad range of tasks,
including interleaved text--image generation~\cite{an2023openleaf}.
These models commonly embed heterogeneous modalities into a shared representation space
and employ a single transformer backbone to capture cross-modal interactions.

Training UMMs typically follows a two-stage paradigm.
In the first stage, models are pretrained to align text and visual representations at scale.
In the second stage, they are adapted to downstream tasks via instruction tuning,
which fine-tunes the model on diverse task-specific instructions paired with expected outputs.
Instruction tuning has become a standard practice for improving alignment with human intent
and enhancing task generalization~\cite{visual_inst_tuning, visual_inst_tuning2}.


Despite these advances, existing UMMs often struggle to generate interleaved text and images
with balanced quality.
In practice, these models tend to exhibit strong performance on language-dominant tasks
while underperforming on vision-dominant tasks, indicating a pronounced modality imbalance
during multimodal learning~\cite{moss, chen2024interleaved}.
This observation raises a natural question:
\emph{can instruction tuning improve image generation quality to the same extent as language modeling, or does modality imbalance remain a fundamental challenge?}

\begin{figure}[bt]
  \centering
  \includegraphics[width=0.7\linewidth]{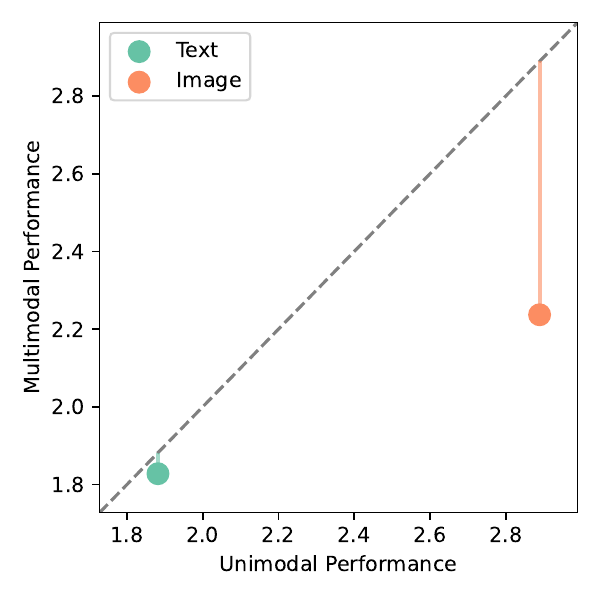}
  \vspace{-1em}
  \caption{Performance gap between unimodal counterparts and the Emu2~\cite{emu2} model after multimodal instruction tuning. Vision-only performance drops substantially more than text-only performance, indicating that image generation is the weaker modality under multimodal optimization.
}
  \label{fig:unimodal}
\end{figure}

To investigate this issue, we conduct two quantitative studies using LoRA-based fine-tuning,
which is widely adopted due to the large scale of modern UMMs.
First, as illustrated in Figure~\ref{fig:unimodal}, we fine-tune unimodal counterparts
(vision-only and text-only models) on the same dataset and compare their performance
against the multimodal model.
We observe that, when comparing unimodal counterparts to the multimodal model,
vision-only performance drops more significantly than text-only performance,
revealing vision as the weaker modality and is more heavily suppressed under multimodal LoRA fine-tuning.

\begin{figure}[bt]
  \centering
  \includegraphics[width=\linewidth]{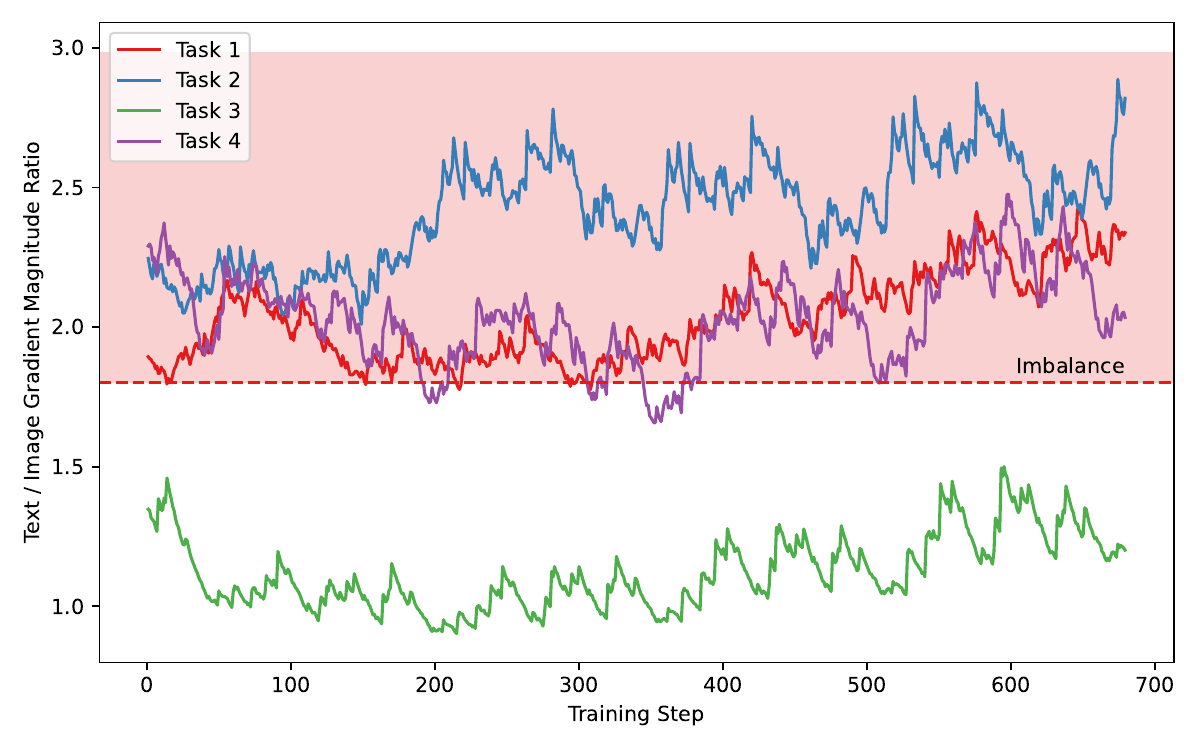}
  \vspace{-1em}
  \caption{Relative gradient magnitude ratio between text and image objectives during multimodal instruction tuning. Curves correspond to Task 1 (image-to-text generation), Task 2 (text-to-image generation), Task 3 (interleaved image–text generation), and Task 4 (question-based interleaved image–text generation). Three out of four tasks exhibit strong text-dominant gradients, revealing severe modality imbalance in LoRA-based instruction tuning.
}
  \vspace{-1em}
  \label{fig:gradient_magnitude}
\end{figure}

Second, we analyze the relative gradient magnitudes of modality-specific losses
with respect to the shared parameters during multimodal instruction tuning.
A large gradient ratio implies that text modality dominates the learning process.
As shown in Figure~\ref{fig:gradient_magnitude}, we find that three out of four evaluated tasks
exhibit significant gradient imbalance. 

To address this challenge, we draw inspiration from balanced multimodal learning
and reformulate modality imbalance as a bi-objective optimization problem.
Each modality is treated as a distinct learning objective, and the goal is to identify
a Pareto-optimal solution that balances optimization across modalities.
Rather than allowing the dominant modality to overwhelm optimization,
we seek a gradient direction that simultaneously benefits all objectives,
guiding training toward a stable trade-off.

Based on this formulation, we propose \emph{Pareto LoRA},
a Pareto-optimal gradient integration method for LoRA-based fine-tuning.
Pareto LoRA ensures that the final update direction aligns with shared descent directions
across modalities, preventing the suppression of weaker modalities
while preserving overall learning efficiency.

We summarize our contributions as follows:
\begin{itemize}
  \item We identify and empirically characterize modality imbalance during LoRA-based fine-tuning of UMMs for interleaved text--image generation.
  \item We propose Pareto LoRA, a Pareto-optimal optimization method that mitigates modality imbalance by balancing modality-specific gradients.
  \item We demonstrate through extensive experiments that Pareto LoRA improves multimodal balance without sacrificing overall performance.
\end{itemize}

\section{Related Work}
\label{sec:related_work}

\paragraph{Unified multimodal models (UMMs).}
Early efforts to unify visual understanding and generation~\cite{emu, emu2, metamorph, illume} often combine large multimodal language models (MLLMs) with diffusion-based decoders, where the diffusion process is conditioned on embeddings produced by an autoregressive language backbone.
While effective for instruction-guided generation, these hybrid designs typically rely on separate objectives and architectural components for language modeling and image synthesis, which may limit the degree of unified multimodal modeling~\cite{geneval}.

More recent approaches aim to treat both multimodal understanding and generation under a unified next-token prediction paradigm~\cite{lwm, vila-u, ugen, chameleon}.
These models differ in how they represent the visual content.
Some~\cite{chameleon, emu3, lavit} employ discrete vision tokenizers (e.g., VQGAN/VQ-VAE~\cite{vqgan, vqvae}) to convert images into sequences of visual tokens, enabling fully autoregressive generation.
Others encode images into continuous latent embeddings for multimodal reasoning, while still relying on diffusion or latent decoders for high-fidelity image synthesis~\cite{illume+, illume}.
Despite architectural differences, most UMMs share an autoregressive transformer backbone for joint multimodal modeling.

In this work, we adopt Emu2~\cite{emu2}, a 37B-parameter unified multimodal model trained with an autoregressive next-element prediction objective over interleaved text and visual embeddings.
Emu2 consists of three components: a visual encoder, an LLM backbone (initialized from LLaMA-33B~\cite{touvron2023llama}), and a diffusion-based visual decoder.
Each input image is encoded into continuous embeddings, projected into the LLM token space, and interleaved with text tokens for multimodal autoregressive modeling.
The predicted visual embeddings are subsequently decoded into images by the visual decoder.
In our experiments, we fine-tune only the LLM backbone and the projection layers using LoRA, while keeping the visual encoder and decoder frozen.

\paragraph{Instruction tuning and parameter-efficient tuning.}
Visual instruction tuning, popularized by LLaVA~\cite{llava}, demonstrates that large language models can be effectively adapted to follow multimodal instructions through supervised fine-tuning on large-scale vision--language data.
Instruction tuning has since been widely applied to multimodal tasks such as image editing~\cite{fuguiding, huang2024smartedit}, controllable image generation~\cite{tian2025mige}, interleaved multimodal understanding~\cite{jiang2024mantis, hu2024instruct}, and interleaved text--image generation~\cite{moss}.

Given the scale of modern MLLMs and UMMs, parameter-efficient fine-tuning (PEFT) methods have become the dominant adaptation strategy.
These methods freeze the majority of pretrained parameters and update only a small set of newly introduced modules~\cite{hu2022lora, sung2022vl}.
LoRA~\cite{hu2022lora} is particularly effective and has been widely adopted for instruction tuning of multimodal foundation models~\cite{weimitigating, gedynamic, moss}.
In this paper, we build upon LoRA and propose a gradient balancing strategy tailored for unified multimodal generation.

\paragraph{Modality imbalance in multimodal learning.}
Recent studies have highlighted modality imbalance in multimodal training, where optimization tends to favor one dominant modality (often text), leading to degraded performance on the weaker modality~\cite{li2023boosting, yang2025learning}.
MMPareto~\cite{wei2024mmpareto} addresses this issue by applying Pareto-based gradient integration between unimodal objectives and multimodal joint learning objectives.

However, in unified multimodal models, text and image generation are tightly coupled within a single autoregressive backbone, making it difficult to cleanly separate unimodal and multimodal losses.
Modality imbalance remains a persistent challenge in modern MLLMs and UMMs~\cite{wu2024commit, zheng2025mllms, zhang2025robust}, often resulting in biased generation behavior toward the dominant modality.

Several methods mitigate imbalance through module-wise dynamic updates, such as D-MoLE~\cite{gedynamic}, which adjusts learning dynamics based on modality difficulty.
In contrast, for unified architectures where both modalities share the same backbone and compete over the same parameters, imbalance must be addressed directly at the gradient level.
We therefore propose \emph{Pareto LoRA}, which explicitly balances text and image generation gradients during LoRA-based instruction tuning.
Unlike MMPareto~\cite{wei2024mmpareto}, which relies on separable unimodal objectives, Pareto LoRA operates directly on coupled modality gradients within a unified autoregressive model, making it particularly suitable for interleaved multimodal generation.

\section{Methodology}
\label{sec:method}

\subsection{Multimodal Instruction Tuning as Multi-Objective Optimization}

When fine-tuning UMMs for interleaved text--image generation
(e.g., recipes or multimodal reasoning tasks),
the training objective naturally consists of two modality-specific components:
text generation and image generation~\cite{show-o, transfusion}.
The unified training objective is commonly written as:
\begin{equation}
  L_{\text{unified}} = L_{\text{text}} + \alpha L_{\text{img}},
\end{equation}
where $\alpha$ is a scalar balancing coefficient.

\begin{figure}[bt]
  \centering
  \includegraphics[width=\linewidth]{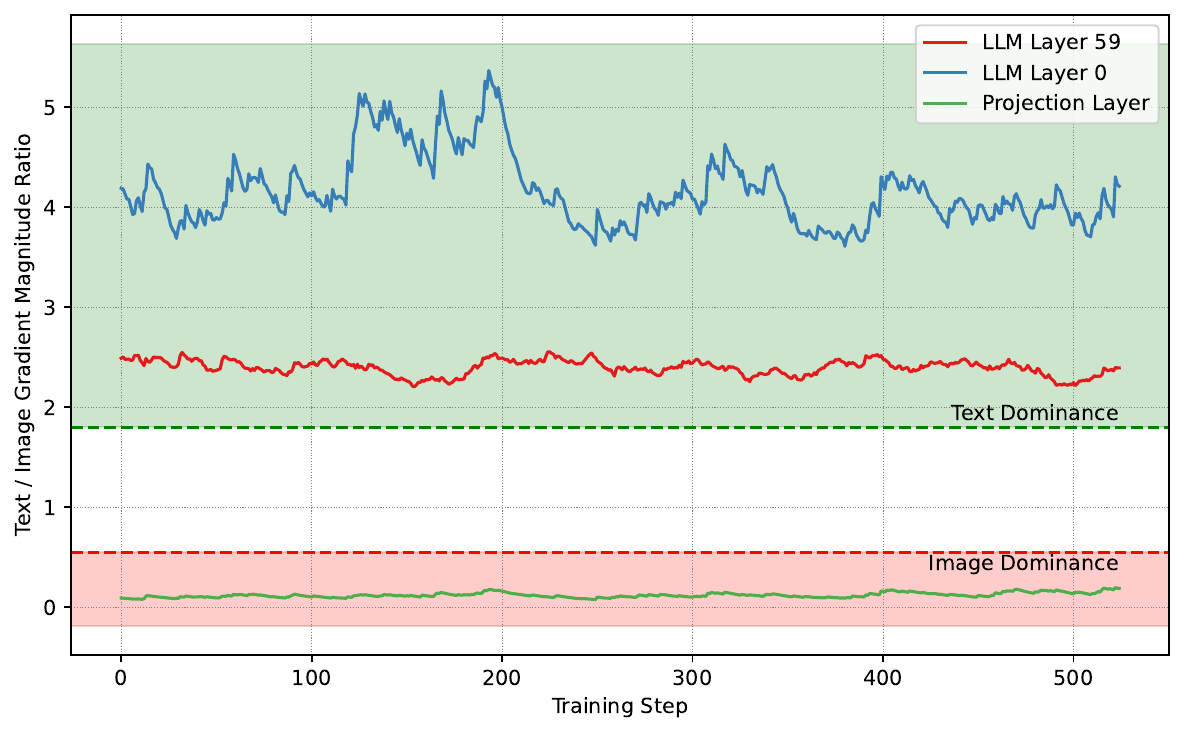}
  \caption{Modality gradient magnitude ratio ($\|g_{\text{txt}}\|_2 / \|g_{\text{img}}\|_2$) of each layer during training. The y--axis measures how much stronger text gradients are than image gradients at that layer. Text gradients dominate most LLM layers, while projection layers exhibit different imbalance behavior, motivating selective gradient modulation.
 }
  \label{fig:rho_layerwise}
  \vspace{-1em}
\end{figure}

While simple and effective in practice, this formulation implicitly assumes that
a fixed loss-level weighting can adequately balance modality learning.
However, as shown in Section~\ref{sec:intro},
optimizing the unified objective often leads to one modality dominating the optimization process,
leaving the other underoptimized.
Although one could tune the coefficient $\alpha$,
we observe that both the \emph{direction} (which modality dominates)
and the \emph{degree} (how severe the imbalance is)
vary across tasks, layers, and training stages
(see Fig~\ref{fig:gradient_magnitude} and Fig~\ref{fig:rho_layerwise}).
Consequently, loss-level reweighting alone is insufficient to address modality imbalance
at the gradient level.

\subsection{Pareto Integration}

Multimodal learning is closely related to multi-task learning,
as both involve jointly optimizing multiple objectives that share parameters~\cite{wei2024mmpareto}.
Motivated by the observed imbalance between modality-specific gradients,
we reformulate multimodal instruction tuning as a bi-objective optimization problem,
where text and image learning are treated as separate but coupled objectives.

Let $g_{\text{txt}}$ and $g_{\text{img}}$ denote the gradients of the text and image losses,
respectively, computed on the same batch $S$ and with respect to the same trainable parameters $m$
(i.e., the LoRA-adapted parameters):
\begin{equation}
  \vspace{-.3em}
  g_{\text{txt}} = \nabla_{m,S} L_{\text{text}}, \quad
  g_{\text{img}} = \nabla_{m,S} L_{\text{img}}.
  \vspace{-.2em}
\end{equation}
Rather than summing these gradients with a fixed coefficient,
we seek a convex combination that yields a descent direction beneficial to both objectives; this leads to the following optimization problem:
\begin{equation}
  \min_{\lambda, \beta}
  \left\| \lambda g_{\text{txt}} + \beta g_{\text{img}} \right\|_2^2
  \quad
  \text{s.t. } \lambda, \beta \ge 0,\;\; \lambda + \beta = 1.
  \label{eq:pareto_opt}
\end{equation}

This formulation corresponds to finding the minimum-norm vector
in the convex hull of $\{g_{\text{txt}}, g_{\text{img}}\}$.
The Multiple Gradient Descent Algorithm (MGDA)~\cite{mgda}
shows that either the minimum-norm solution is zero, indicating a Pareto-stationary point
(a necessary condition for Pareto optimality), 
or the solution provides a common descent direction
that simultaneously improves all objectives.

Following~\cite{sener2018multi},
the optimal weighting coefficient admits a closed-form solution:
\begin{equation}
  \lambda =
  \mathrm{clip}
  \left(
    \frac{(g_{\text{txt}} - g_{\text{img}})^\top g_{\text{txt}}}
         {\| g_{\text{img}} - g_{\text{txt}} \|_2^2},
    0, 1
  \right),
  \quad
  \beta = 1 - \lambda.
  \label{eq:pareto_sol}
\end{equation}

\subsection{Pareto LoRA}

\begin{figure}[t]
  \centering
  \includegraphics[width=\linewidth]{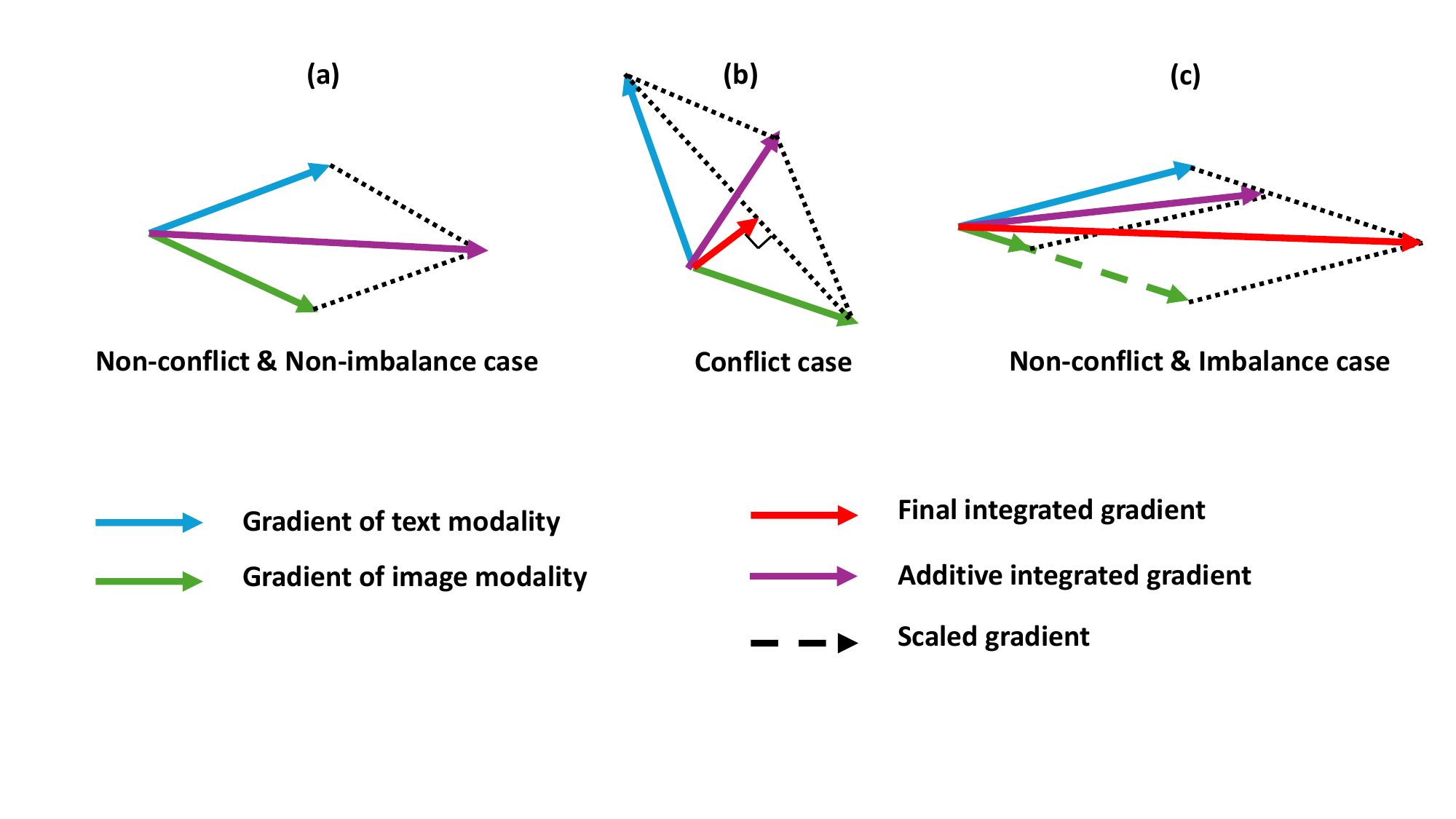}
  \vspace{-4em}
  \caption{Overview of Pareto LoRA gradient modulation. Depending on gradient cosine similarity and magnitude imbalance, Pareto LoRA applies (i) Pareto integration under conflict, (ii) gradient rescaling under severe imbalance, or (iii) no modification when gradients are aligned and balanced.
}
  \vspace{-1em}
  \label{fig:method}
\end{figure}

Based on the above formulation, we propose \emph{Pareto LoRA}, a gradient integration strategy tailored for LoRA-based multimodal fine-tuning. Rather than applying Pareto optimization uniformly at every step, Pareto LoRA first diagnoses the relationship between modality-specific gradients using two metrics: (i) cosine similarity, which captures directional conflict, and (ii) gradient magnitude ratio, which measures imbalance strength.

Let $g_{\text{txt}}=\nabla_{\mathbf{m}}L_{\text{txt}}$
and $g_{\text{img}}=\nabla_{\mathbf{m}}L_{\text{img}}$
denote the text and image gradients with respect to the trainable LoRA parameters $\mathbf{m}$.
We compute:
\begin{equation}
  \vspace{-.4em}
  c = \frac{g_{\text{txt}}^\top g_{\text{img}}}
  {\|g_{\text{txt}}\|_2\,\|g_{\text{img}}\|_2},
  \qquad
  r = \frac{\|g_{\text{txt}}\|_2}{\|g_{\text{img}}\|_2}.
\end{equation}

Figure~\ref{fig:method} illustrates three cases and their corresponding strategies:

\paragraph{\textbf{Conflict case ((b) in Fig~\ref{fig:method}).}}
We first consider the case where $c < 0$. In this case, the text and image gradients are directionally conflicting. It is essential to find a descent direction that is shared by both objectives. Therefore, we solve the Pareto optimization problem in Equation~\ref{eq:pareto_opt}, obtaining $\lambda$ and $\beta$ using Equation~\ref{eq:pareto_sol}, which yields a non-conflicting update direction. The final gradient is then given by the corresponding weighted combination: $g \leftarrow \lambda g_{\text{txt}} + \beta g_{\text{img}}$.

\paragraph{\textbf{Non-conflict but imbalanced case ((c) in Fig~\ref{fig:method}).}}
For the case where $c \ge 0$ and $r>\tau$, the gradients are directionally aligned, and any convex combination provides a common descent direction. However, significant imbalance in gradient strength can still cause one modality to dominate learning. To mitigate this, we rescale the weaker gradient to match the stronger one before aggregation. This avoids overly shrinking the dominant modality gradient, which could otherwise reduce effective learning. 
Specifically, letting $g_{\text{weak}}$ and $g_{\text{strong}}$ denote the gradients
with smaller and larger norms, respectively, we apply
\begin{equation}
  \vspace{-.3em}
  g_{\text{weak}} \leftarrow
  g_{\text{weak}}
  \cdot
  \frac{\|g_{\text{strong}}\|_2}{\|g_{\text{weak}}\|_2}.
  \vspace{-.3em}
\end{equation}
The final update is then computed by standard summation:
\begin{equation}
  \vspace{-.3em}
  g \leftarrow g_{\text{txt}} + g_{\text{img}}.
  \vspace{-.3em}
\end{equation}

\paragraph{\textbf{Non-conflict and balanced case((a) in Fig~\ref{fig:method}).}}
When gradients are aligned and comparable in magnitude ($c\ge 0$ and $r\le\tau$), we apply no modification and use the vanilla multimodal gradient update.

Algorithm~\ref{alg:palora} summarizes Pareto LoRA.

\begin{algorithm}[t]
\caption{Pareto LoRA gradient modulation. The update adaptively switches between Pareto-optimal integration and magnitude matching based on gradient conflict and imbalance diagnostics.
}
\label{alg:palora}
\begin{algorithmic}[1]
\Require Text loss $L_{\text{txt}}$, image loss $L_{\text{img}}$, threshold $\tau$
\State Compute gradients
$g_{\text{txt}}=\nabla_{\mathbf{m}}L_{\text{txt}}$,
$g_{\text{img}}=\nabla_{\mathbf{m}}L_{\text{img}}$
\State Compute cosine similarity
$
c=\frac{g_{\text{txt}}^\top g_{\text{img}}}
{\|g_{\text{txt}}\|_2\|g_{\text{img}}\|_2}
$
\State Compute magnitude ratio
$
r=\frac{\|g_{\text{txt}}\|_2}{\|g_{\text{img}}\|_2}
$

\If{$c<0$} \Comment{Conflict case}
    \State Solve Eq.~\ref{eq:pareto_opt} to obtain $(\lambda,\beta)$
    \State $g \gets \lambda g_{\text{txt}}+\beta g_{\text{img}}$
\ElsIf{$c\ge 0$ \textbf{and} $r>\tau$} \Comment{Imbalance case}
    \State Rescale weaker gradient to match the stronger norm
    \State $g \gets g_{\text{txt}}+g_{\text{img}}$
\Else \Comment{Balanced case}
    \State $g \gets g_{\text{txt}}+g_{\text{img}}$
\EndIf

\State Update LoRA parameters using gradient $g$
\end{algorithmic}
\end{algorithm}

\begin{figure}[t]
  \centering
  \includegraphics[width=\linewidth]{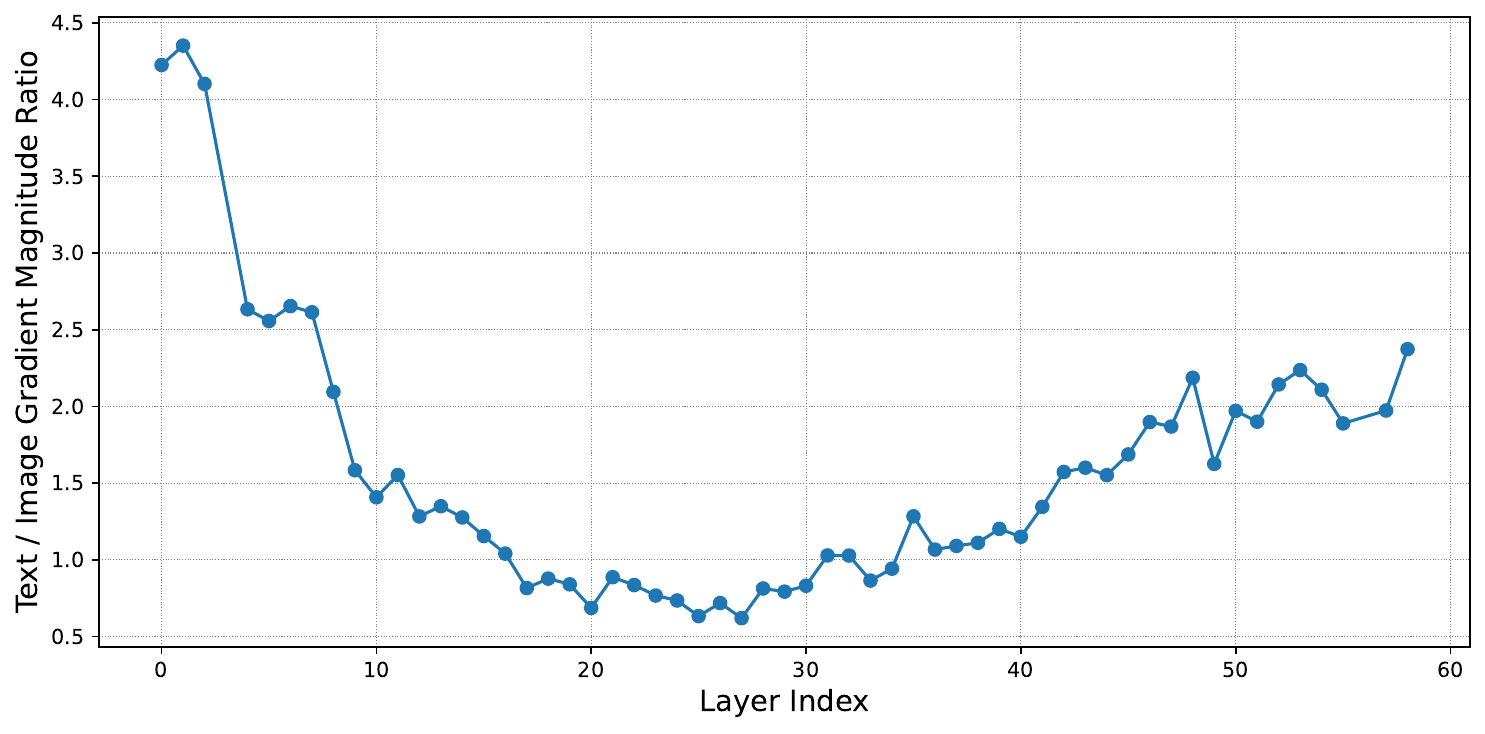}
  \caption{Per-layer modality gradient magnitude ratio across the 60 LLM layers of Emu2 model~\cite{emu2} on CoMM~\cite{chen2025comm} Task 4. Text gradients dominate strongly in lower layers, become more balanced in intermediate layers, and exhibit renewed text dominance in upper layers closer to generation outputs. This non-uniform depth-wise imbalance motivates group-wise and selective modulation rather than uniform gradient balancing across all layers.
}
  \label{fig:avg_perlayer}
  \vspace{-1em}
\end{figure}

In addition to gradient direction and magnitude modulation, it is important to determine \emph{which layers} to modulate, since imbalance scenarios differ across layers in UMMs (Figure~\ref{fig:rho_layerwise}). According to our observations, LLM layers are typically dominated by the text modality, while the projection layer is dominated by the image modality. As shown in Figure~\ref{fig:avg_perlayer}, lower LLM layers (closer to the input) are more text-dominated than upper layers, while middle layers tend to be more modality-balanced. This motivates a layer-granular modulation strategy. However, modern UMMs are extremely large (e.g., 60 LLM layers in Emu2~\cite{emu2}), making fully layer-wise computation of Equation~\ref{eq:pareto_sol} computationally expensive. We therefore group neighboring layers and apply modulation at the \textit{group level}, which is both efficient and reasonable given the similarity of imbalance patterns across adjacent layers. Moreover, since middle layers are often balanced, \textit{selective modulation} is preferred. In our experiments, modulating lower layers achieves the best performance, and we provide an ablation study on targeted modulation layers in Section~\ref{sec:ablation_target_params}.

By selectively invoking Pareto integration only when necessary, Pareto LoRA achieves balanced multimodal optimization while preserving training stability and efficiency.
\section{Experiments}
\label{sec:experiment}

\begin{table*}[t]
\caption{Main quantitative results on CoMM Task 1 (image-to-text) and Task 2 (text-to-image). Pareto LoRA consistently improves perceptual quality and image coherence over vanilla LoRA. Relative gains (\%) are reported w.r.t.\ vanilla LoRA. Best results within each task are highlighted in \textbf{bold} and second best results are highlighted \underline{underlined}.}
\label{tab:task12_main}
\centering
\setlength{\tabcolsep}{6pt}
\begin{tabular}{l|c|cc}
\toprule
& \textbf{Task 1} 
& \multicolumn{2}{c}{\textbf{Task 2}} \\
\cmidrule(lr){2-2}
\cmidrule(lr){3-4}
Method & Text quality ($\uparrow$) & Perceptual Quality ($\uparrow$) & Image Coherence ($\uparrow$) \\
\midrule
Zero Shot & 1.05 & 1.02 & 1.02 \\
Vanilla LoRA & \underline{1.22} & 1.67 & 1.59 \\
GradNorm & 1.14 & 1.70 & 1.54 \\
Step Balance & 1.02 & 1.68 & 1.45 \\
Pareto LoRA & \textbf{1.24} (\textcolor{Green}{+1.64\%}) & 2.12 (\textcolor{Green}{+26.95\%}) & 1.94 (\textcolor{Green}{+22.01\%}) \\
\midrule
Unimodal Upper Bound & 1.40 & 2.06 & 1.91 \\
\bottomrule
\end{tabular}
\end{table*}

\begin{table*}[t]
    \caption{Results on CoMM Task 3 (interleaved multimodal content generation). Pareto LoRA improves both image quality and overall helpfulness compared to vanilla LoRA. Best results are highlighted in \textbf{bold} and second best results are highlighted \underline{underlined}.
}
    \label{tab:task3}
    \centering
    \begin{tabular}{lccccc}
        \toprule
        Method & Text quality ($\uparrow$) & Perceptual quality ($\uparrow$) & Image Coherence ($\uparrow$) & TIC ($\uparrow$) & Helpfulness ($\uparrow$) \\
        \midrule
        Zero Shot & 1.32 & 1.37 & 1.30 & 1.34 & 1.91 \\
        Vanilla LoRA & \underline{1.83} & \underline{2.16} & \underline{2.24} & \textbf{2.00} & \underline{2.09} \\
        GradNorm & 1.60 & 1.88 & 1.74 & 1.65 & 1.87 \\
        Step Balance & 1.46 & 1.73 & 1.76 & 1.63 & 2.02 \\
        Pareto LoRA & \textbf{1.97} (\textcolor{Green}{+7.65\%}) & \textbf{2.79} (\textcolor{Green}{+29.17\%}) & \textbf{2.30} (\textcolor{Green}{+2.68\%}) & \underline{1.94} (\textcolor{Red}{-3.00\%}) & \textbf{2.20} (\textcolor{Green}{+5.26\%}) \\
        \midrule
        Unimodal Upper Bound & 1.88 & 2.82 & 2.89 & 2.44 & 2.94 \\
        \bottomrule
    \end{tabular}
    \vspace{-0.5em}
\end{table*}

\begin{table*}[t]
    \caption{Results on CoMM Task 4 (question-based interleaved generation). Pareto LoRA substantially improves image-related metrics but may slightly reduce text helpfulness in language-centric regimes, highlighting task-dependent modulation effects. Best results within each task are highlighted in \textbf{bold} and second best results are highlighted \underline{underlined}.
}
    \label{tab:task4}
    \centering
    \begin{tabular}{lccccc}
        \toprule
        Method & Text quality ($\uparrow$) & Perceptual quality ($\uparrow$) & Image Coherence ($\uparrow$) & TIC ($\uparrow$) & Helpfulness ($\uparrow$) \\
        \midrule
        Zero Shot & 1.35 & 1.36 & 1.25 & 1.30 & \textbf{1.85} \\
        Vanilla LoRA & \textbf{1.70} & \underline{1.96} & \underline{1.96} & 1.72 & \underline{1.83} \\
        GradNorm & 1.51 & 1.33 & 1.15 & \underline{1.87} & \underline{1.83} \\
        Step Balance & \underline{1.60} & 1.54 & 1.60 & 1.35 & 1.79 \\
        Pareto LoRA & 1.57 (\textcolor{Red}{-7.65\%}) & \textbf{2.84} (\textcolor{Green}{+44.90\%}) & \textbf{2.54} (\textcolor{Green}{+29.59\%}) & \textbf{2.02} (\textcolor{Green}{+17.44\%}) & 1.66 (\textcolor{Red}{-9.29\%}) \\
        \midrule
        Unimodal Upper Bound & 1.62 & 2.10 & 2.04 & 1.99 & 1.81 \\
        \bottomrule
    \end{tabular}
\end{table*}

\begin{table*}[t]
\caption{Ablation study on targeted gradient modulation layers across the four CoMM tasks. Applying Pareto LoRA to lower LLM layer groups yields the strongest gains on visually grounded tasks, while upper-layer modulation can benefit language-centric settings such as Task 4. Best results within each task are highlighted in bold.}
\label{tab:ablation}
\centering
\resizebox{\textwidth}{!}{
\begin{tabular}{l|c|cc|ccccc|ccccc}
\toprule
& \textbf{Task 1}  & \multicolumn{2}{c|}{\textbf{Task 2}} & \multicolumn{5}{c|}{\textbf{Task 3}} & \multicolumn{5}{c}{\textbf{Task 4}} \\
\cmidrule(lr){2-2} \cmidrule(lr){3-4} \cmidrule(lr){5-9} \cmidrule(lr){10-14}
Targeted Group & TextQ & PercQ & ImgC & TextQ & PercQ & ImgC & TIC & Help & TextQ & PercQ & ImgC & TIC & Help \\
\midrule
Zero Shot & 1.05 & 1.02 & 1.02 & 1.32 & 1.37 & 1.30 & 1.34 & 1.91 & 1.35 & 1.36 & 1.25 & 1.30 & 1.85 \\
\midrule
Lower & 1.24 & \textbf{2.12} & \textbf{1.94} & \textbf{1.97} & \textbf{2.79} & \textbf{2.30} & \textbf{1.94} & \textbf{2.20} & 1.57 & \textbf{2.84} & \textbf{2.54} & \textbf{2.02} & 1.66 \\
Middle & \textbf{1.31} & 1.84 & 1.58 & 1.52 & 1.96 & 1.83 & 1.63 & 1.74 & 1.66 & 2.43 & 2.29 & 1.96 & 1.83 \\
Upper & 1.28 & 1.56 & 1.39 & 1.48 & 2.02 & 1.91 & 1.70 & 1.87 & \textbf{1.77} & 2.28 & 2.13 & 1.83 & \textbf{1.87} \\
\bottomrule
\end{tabular}
}
\footnotesize{
TextQ: Text quality, PercQ: Perceptual quality, ImgC: Image coherence, TIC: Text--image coherence, Help: Helpfulness.
}
\end{table*}

\begin{table*}[t]
\caption{Ablation on Pareto LoRA modulation components across the four CoMM tasks. Handling both gradient conflict (Pareto-only) and magnitude imbalance (Rescale-only) is necessary for consistent multimodal improvements, with the full method performing best overall. Best results within each task are highlighted in \textbf{bold} and second best results are highlighted \underline{underlined}.}
\label{tab:ablation_component}
\centering
\resizebox{\textwidth}{!}{
\begin{tabular}{l|c|cc|ccccc|ccccc}
\toprule
& \textbf{Task 1}  & \multicolumn{2}{c|}{\textbf{Task 2}} & \multicolumn{5}{c|}{\textbf{Task 3}} & \multicolumn{5}{c}{\textbf{Task 4}} \\
\cmidrule(lr){2-2} \cmidrule(lr){3-4} \cmidrule(lr){5-9} \cmidrule(lr){10-14}
Variant & TextQ & PercQ & ImgC & TextQ & PercQ & ImgC & TIC & Help & TextQ & PercQ & ImgC & TIC & Help \\
\midrule
Vanilla LoRA & 1.22 & 1.67 & \underline{1.59} & \underline{1.83} & 2.16 & 2.24 & \textbf{2.00} & 2.09 & 1.70 & 1.96 & 1.96 & 1.72 & 1.83 \\
Rescale-only & 1.20 & \underline{1.85} & 1.56 & 1.62 & \underline{2.69} & \textbf{2.34} & \underline{1.99} & \underline{2.16} & \underline{1.67} & \underline{2.37} & \underline{2.36} & \underline{1.95} & \underline{1.90} \\
Pareto-only  & \underline{1.23} & 1.68 & 1.50 & 1.64 & 2.17 & 1.87 & 1.73 & 2.00 & \textbf{1.79} & 1.84 & 1.98 & 1.64 & \textbf{1.91} \\
Full Pareto LoRA & \textbf{1.24} & \textbf{2.12} & \textbf{1.94} & \textbf{1.97} & \textbf{2.79} & \underline{2.30} & 1.94 & \textbf{2.20} & 1.57 & \textbf{2.84} & \textbf{2.54} & \textbf{2.02} & 1.66 \\
\bottomrule
\end{tabular}
}
\footnotesize{
TextQ: Text quality, PercQ: Perceptual quality, ImgC: Image coherence, TIC: Text--image coherence, Help: Helpfulness.
}

\end{table*}

\begin{table*}[t]
\caption{Ablation on $\tau$ values. Best results within each task are highlighted in \textbf{bold} and second best results are highlighted \underline{underlined}.}
\label{tab:ablation_tau}
\centering
\resizebox{\textwidth}{!}{
\begin{tabular}{l|c|cc|ccccc|ccccc}
\toprule
& \textbf{Task 1}  & \multicolumn{2}{c|}{\textbf{Task 2}} & \multicolumn{5}{c|}{\textbf{Task 3}} & \multicolumn{5}{c}{\textbf{Task 4}} \\
\cmidrule(lr){2-2} \cmidrule(lr){3-4} \cmidrule(lr){5-9} \cmidrule(lr){10-14}
$\tau$ & TextQ & PercQ & ImgC & TextQ & PercQ & ImgC & TIC & Help & TextQ & PercQ & ImgC & TIC & Help \\
\midrule
0.5 & 1.24 & \textbf{2.12} & \textbf{1.94} & \textbf{1.97} & \textbf{2.79} & \underline{2.30} & 1.94 & 2.20 & \underline{1.57} & \textbf{2.84} & \textbf{2.54} & 2.02 & 1.66 \\
0.6 & \textbf{1.40} & \underline{1.99} & 1.84 & \underline{1.93} & 2.68 & 2.19 & 1.93 & \underline{2.26} & 1.54 & 2.57 & \underline{2.48} & 1.97 & \textbf{1.84} \\
0.7 & \textbf{1.40} & 1.88 & 1.85 & 1.91 & \underline{2.69} & \textbf{2.38} & \underline{2.03} & \textbf{2.39} & 1.56 & 2.62 & \textbf{2.54} & \textbf{2.09} & \underline{1.74} \\
0.8 & \underline{1.36} & 1.86 & \underline{1.89} & 1.81 & \underline{2.76} & 2.28 & \textbf{2.14} & 2.16 & \textbf{1.60} & \underline{2.63} & 2.46 & \underline{2.04} & \underline{1.74} \\
\bottomrule
\end{tabular}
}
\footnotesize{
TextQ: Text quality, PercQ: Perceptual quality, ImgC: Image coherence, TIC: Text--image coherence, Help: Helpfulness.
}
\vspace{-1em}
\end{table*}

\subsection{Datasets and Experimental Setup}

We conduct experiments on the CoMM dataset~\cite{chen2025comm}, which consists of four multimodal instruction tuning tasks:
Task~1 (image-to-text generation),
Task~2 (text-to-image generation),
Task~3 (interleaved image--text content generation),
and Task~4 (question-based interleaved image--text generation).
We follow the published CoMM train/evaluation splits for all experiments.

Our evaluation protocol follows~\cite{liu2024holistic}.
Specifically, we adopt GPT-4o as an automatic multimodal judge to assess model outputs from five aspects:
\emph{text quality}, \emph{perceptual quality}, \emph{image coherence}, \emph{text--image coherence (TIC)}, and \emph{helpfulness}.
For each aspect, GPT-4o assigns a discrete score from $\{1,2,3,4,5\}$ according to the detailed criteria provided in Table~8 of~\cite{liu2024holistic},
where 1 denotes the worst quality (or empty output) and 5 denotes the best quality.
In addition, we instruct GPT-4o to provide brief explanations to improve interpretability.

We fine-tune the Emu2 model~\cite{emu2} using LoRA with rank $r{=}32$, applied to all linear layers in the LLM backbone.
We use a learning rate of $1\times 10^{-5}$ with BF16 precision.
Training is performed on 8 AMD Instinct™ MI300X GPUs with a per-GPU batch size of 1.
We set $\tau=0.5$ in all experiments unless otherwise stated. We also provide an ablation study on values of $\tau$ in Section~\ref{sec:ablation_tau}. 

We compare our proposed Pareto LoRA against vanilla LoRA, which applies standard LoRA fine-tuning without gradient modulation. 
We also include loss-level reweighting methods: GradNorm~\cite{chen2018gradnorm}, which reweights losses by matching gradient magnitudes, and Step Balance~\cite{guo2025m2}. without fine-tuning.
Finally, we include a unimodal counterpart by fine-tuning the model with only a single objective (text-only or image-only).
We report the best score per metric among these unimodal models, representing the performance ceiling when no cross-modal trade-off is required.

\subsection{Instruction tuning on CoMM dataset}

\paragraph{Quantitative results.}
Tables~\ref{tab:task12_main},~\ref{tab:task3},~\ref{tab:task4} report our main results on the four CoMM tasks in comparison to the baseline method (vanilla LoRA).
Overall, Pareto LoRA consistently improves multimodal instruction tuning performance, outperforming vanilla LoRA by up to 44.90\% in perceptual image generation quality while also achieving slightly better text generation results.
These gains suggest that by modulating gradient directions and relative modality strength, Pareto LoRA mitigates the dominance of the text modality during tuning of LLM layers (which constitute the majority of model parameters), thereby enhancing image modality learning without sacrificing language capability.
In contrast, conventional loss balancing methods (GradNorm~\cite{chen2018gradnorm}, Step Balance~\cite{guo2025m2}) struggle to address cross-modal gradient conflicts and heterogeneous layer-wise modality imbalance, often resulting in unstable trade-offs or degraded performance.

However, we observe a performance drop in Task~4 on the text quality metric and overall helpfulness as in Table~\ref{tab:task4}.
We attribute this to the distinct input distribution of Task~4, where the context consists of open-ended questions in pure text with little or no visual grounding, whereas Task~3 contains both textual instructions and image inputs.
In such a language-centric regime, multimodal gradient balancing may become less beneficial and can slightly over-regularize the optimization of text objectives, leading to reduced helpfulness.
This highlights that Pareto LoRA is most effective in tasks with strong multimodal interactions and genuine cross-modal gradient competition.



\begin{figure}[t]
  \centering
  \includegraphics[width=\linewidth]{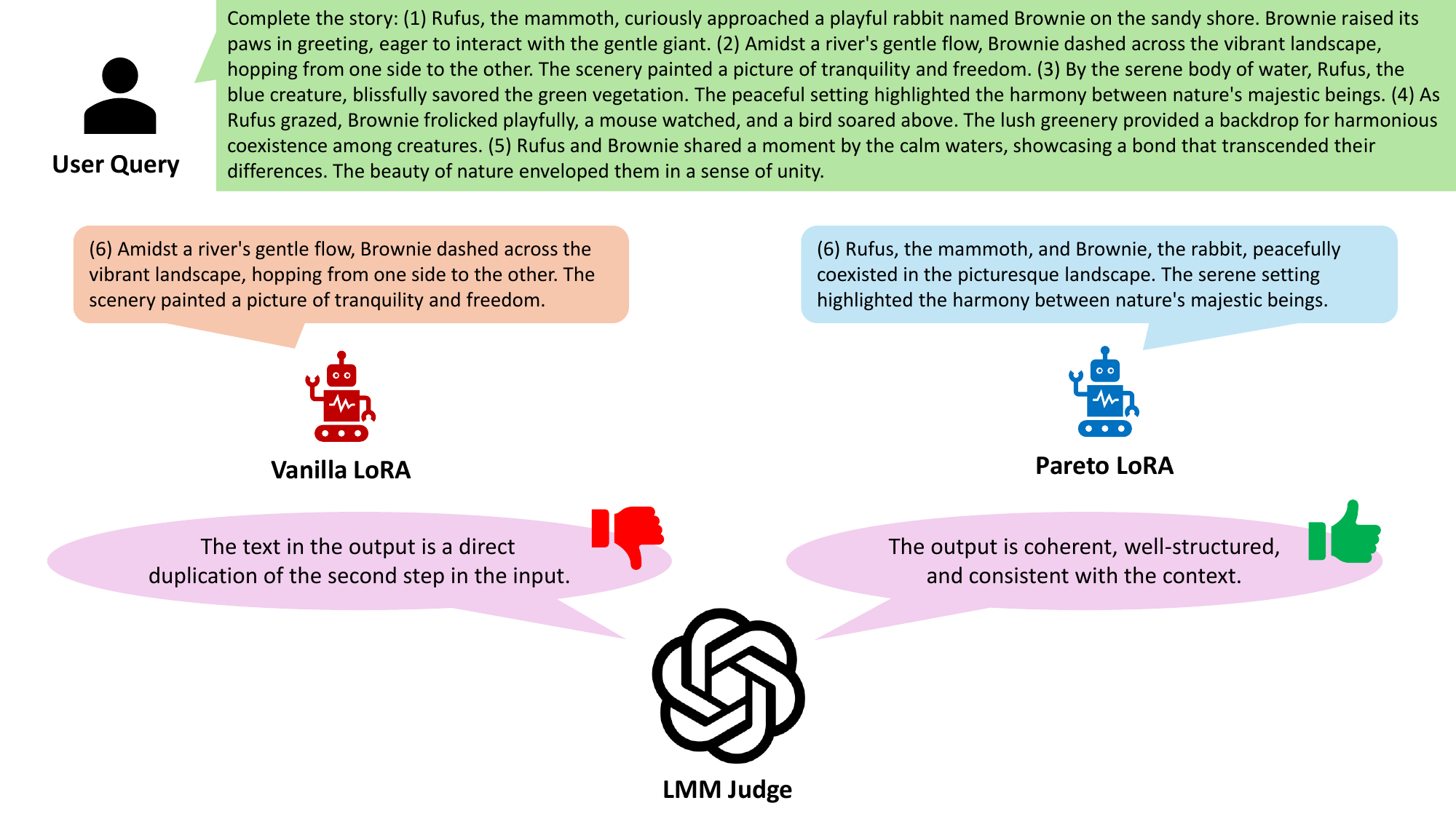}
  \caption{Qualitative comparison of text generation. Vanilla LoRA exhibits degeneration with duplicated content copied from the user query, whereas Pareto LoRA produces more concise and coherent responses.
}
  \vspace{-1em}
  \label{fig:qual_task1}
\end{figure}

\begin{figure}[t]
  \centering
  \includegraphics[width=\linewidth]{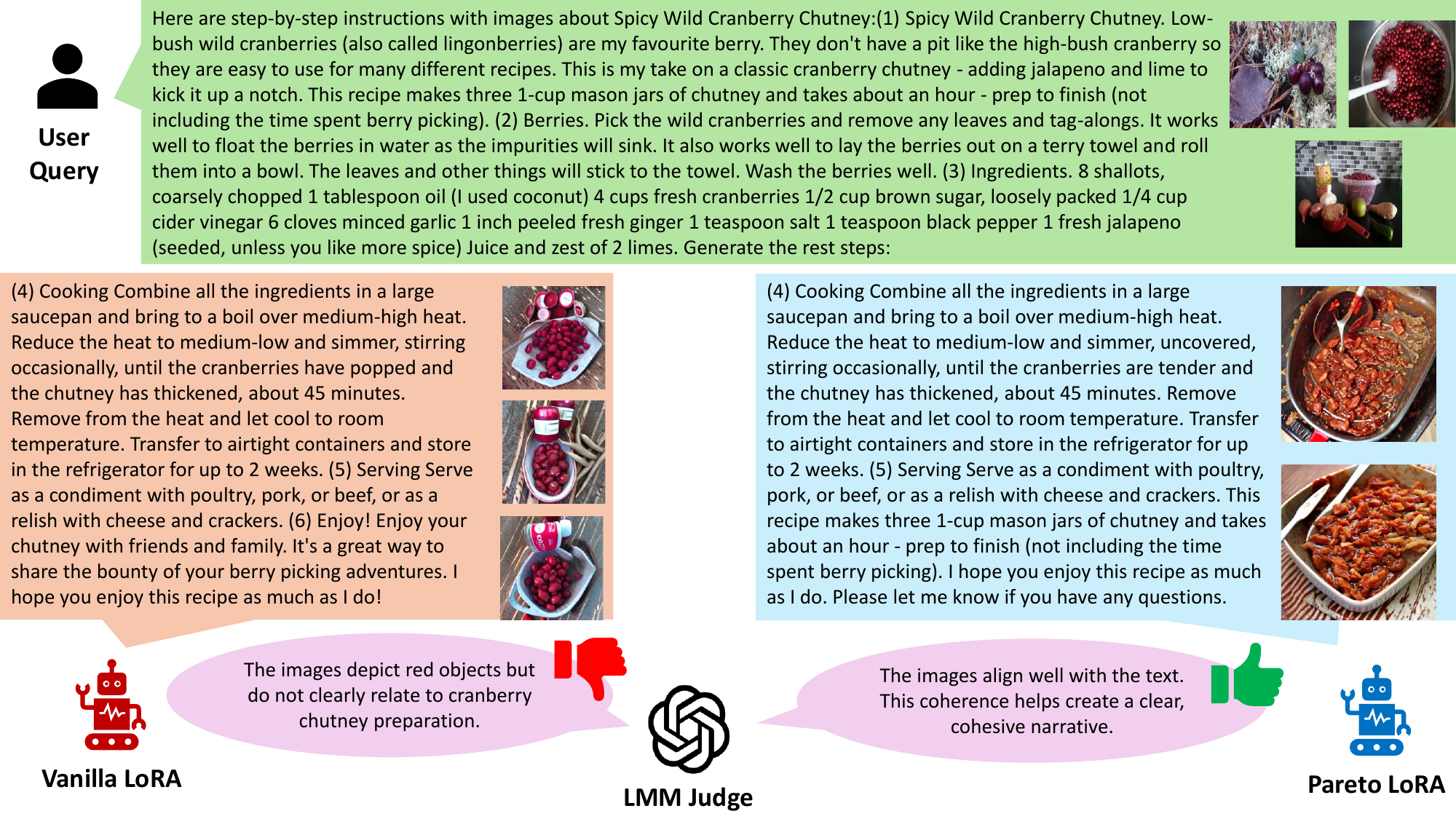}
  \caption{Qualitative comparison of interleaved text--image generation on CoMM. Vanilla LoRA often produces images that are weakly grounded or visually distorted despite fluent text, whereas Pareto LoRA improves perceptual quality and text--image coherence, yielding more consistent multimodal narratives.
}
  \vspace{-1em}
  \label{fig:qual_task3}
\end{figure}

\paragraph{Qualitative results}
Figures~\ref{fig:qual_task1}, ~\ref{fig:qual_task3} and ~\ref{fig:qual_task4} present qualitative comparisons between our Pareto LoRA and vanilla LoRA.
Overall, the visual examples demonstrate that Pareto LoRA produces more coherent and higher-quality multimodal generations.
In particular, vanilla LoRA occasionally exhibits repetitive text outputs that echo the user query, whereas Pareto LoRA generates more fluent and diverse responses. 
Moreover, Pareto LoRA improves perceptual image quality, avoiding the distortion and blurriness observed in vanilla LoRA.
Finally, Pareto LoRA yields stronger text--image coherence in interleaved generation, producing less repetitive and more consistent multimodal content. 

\subsection{Ablation on Targeted Parameters}
\label{sec:ablation_target_params}

We conduct an ablation study to examine the effect of applying Pareto gradient modulation to different subsets of LoRA-injected layers.
As shown in Figure~\ref{fig:rho_layerwise}, modality imbalance varies across the LLM backbone: although all layers exhibit text-dominant gradients, the degree of imbalance is not uniform.

To study targeted modulation, we partition the 60 LLM layers into five consecutive groups.
The \emph{lower} group is closest to the input, the \emph{upper} group is closest to the output, and the remaining groups span the intermediate layers.
We then apply gradient modulation to only one group at a time and evaluate performance.

Table~\ref{tab:ablation} reports the results.
Overall, we find that modulating the \emph{lower} layers yields the most consistent improvements on visually grounded generation tasks (Tasks~2--3), which motivates our default choice in the main experiments.
This suggests that earlier layers play an important role in integrating visual representations, and modulation at this stage can better alleviate text dominance during multimodal alignment.

Interestingly, we observe that modulating upper layers can be beneficial in more language-centric settings.
For example, in Task~4, modulating the \emph{upper} group achieves the best text quality and helpfulness.
This is expected since Task~4 primarily involves open-ended text questions with minimal visual grounding, and the upper layers are more directly responsible for high-level semantic reasoning and final response generation.

These results indicate that the optimal modulation layer group depends on the modality composition of the downstream task, while lower-layer modulation provides the strongest overall gains for multimodal generation.

\subsection{Ablation on Modulation Components}
To isolate the contribution of each component in Pareto LoRA,
we evaluate two simplified variants:
(i) \emph{Pareto-only}, which applies MGDA integration only in the conflict case,
and (ii) \emph{Rescale-only}, which performs gradient magnitude matching without Pareto optimization.

Table~\ref{tab:ablation_component} shows that the full Pareto LoRA achieves the most consistent improvements across all four CoMM tasks,
indicating that jointly handling gradient conflict and magnitude imbalance is beneficial for balanced multimodal instruction tuning.
Among the simplified variants, \emph{Rescale-only} typically yields the second-best performance, especially on image-related metrics,
suggesting that gradient strength imbalance is a primary driver of modality suppression.
This observation is consistent with our earlier analysis of modality gradient magnitude disparities
(Figure~\ref{fig:gradient_magnitude}).

\subsection{Ablation on $\tau$}
\label{sec:ablation_tau}

We present the results of ablation on the ratio threshold $\tau$. As shown in Table~\ref{tab:ablation_tau}, when $\tau$ is larger, text-related metrics (e.g. TextQ and helpfulness) are slightly improved, while $\tau=0.5$ achieves stronger perceptual and image coherence performance. We adopt $\tau=0.5$ as it offers the best overall balance in our setting.
\section{Conclusion}
\label{sec:conclusion}

Unified multimodal models enable interleaved text-image understanding and generation within a single autoregressive framework, yet they suffer from severe modality imbalance during instruction tuning. Our empirical analysis shows that LoRA-based fine-tuning is dominated by text gradients, suppressing vision learning and limiting multimodal generation quality.
To address this, we propose Pareto LoRA, a Pareto-optimal gradient integration strategy that balances modality-specific objectives by modulating gradient direction and strength. Pareto LoRA activates selectively when gradient conflict or imbalance is detected, maintaining training stability while preventing weaker modalities from being overwhelmed.
Experiments on the CoMM benchmark with Emu2 show that Pareto LoRA significantly improves perceptual image quality and multimodal coherence over vanilla LoRA, with gains up to 44.9\%. Overall, Pareto LoRA offers a simple and effective solution to modality imbalance in unified multimodal instruction tuning.

{
    \small
    \bibliographystyle{ieeenat_fullname}
    \bibliography{main}
}

\end{document}